\documentclass[conference]{IEEEtran}
\IEEEoverridecommandlockouts
\usepackage{cite}
\usepackage{amsmath,amssymb,amsfonts}
\usepackage{algorithmic}
\usepackage{algorithm}
\usepackage{graphicx}
\usepackage{textcomp}
\usepackage{booktabs}
\usepackage{xcolor}
\usepackage{diagbox}
\usepackage{colortbl}
\usepackage{adjustbox}

\newcommand{\applyColor}[1]{
  \ifnum#1<60 \cellcolor{blue!40}#1\%
  \else\ifnum#1<70 \cellcolor{blue!20}#1\%
  \else\ifnum#1<80 \cellcolor{yellow!20}#1\%
  \else\ifnum#1<90 \cellcolor{yellow!40}#1\%
  \else\cellcolor{red!40}#1\%
  \fi\fi\fi\fi
}

\def\BibTeX{{\rm B\kern-.05em{\sc i\kern-.025em b}\kern-.08em
    T\kern-.1667em\lower.7ex\hbox{E}\kern-.125emX}}
\begin{document}

\title{Enhanced Anomaly Detection in Automotive Systems Using SAAD: Statistical Aggregated Anomaly Detection}

\author{\IEEEauthorblockN{1\textsuperscript{st} Dacian Goina}
\IEEEauthorblockA{\textit{Faculty of Mathematics} \\ \textit{and Computer Science} \\
\textit{West University of Timisoara}\\
Timisoara, Timis \\
dacian.goina00@e-uvt.ro}
\and
\IEEEauthorblockN{2\textsuperscript{nd} Eduard Hogea}
\IEEEauthorblockA{\textit{Faculty of Mathematics} \\ \textit{and Computer Science} \\
\textit{West University of Timisoara}\\
Timisoara, Timis \\
eduard.hogea00@e-uvt.ro}

\and
\IEEEauthorblockN{3\textsuperscript{rd} George Maties}
\IEEEauthorblockA{\textit{Faculty of Electrical Engineering} \\
\textit{Gheorghe Asachi} \\ \textit{Technical University of Iasi}\\
Iasi, Iasi\\
george.maties@gmail.com}
}

\maketitle

\begin{abstract}
This paper presents a novel anomaly detection methodology termed Statistical Aggregated Anomaly Detection (SAAD). The SAAD approach integrates advanced statistical techniques with machine learning, and its efficacy is demonstrated through validation on real sensor data from a Hardware-in-the-Loop (HIL) environment within the automotive domain. The key innovation of SAAD lies in its ability to significantly enhance the accuracy and robustness of anomaly detection when combined with Fully Connected Networks (FCNs) augmented by dropout layers. Comprehensive experimental evaluations indicate that the standalone statistical method achieves an accuracy of 72.1\%, whereas the deep learning model alone attains an accuracy of 71.5\%. In contrast, the aggregated method achieves a superior accuracy of 88.3\% and an F1 score of 0.921, thereby outperforming the individual models. These results underscore the effectiveness of SAAD, demonstrating its potential for broad application in various domains, including automotive systems.

\end{abstract}


\section{Introduction}

\begin{figure*}[t]
    \centering
    \includegraphics[width = 1\textwidth]{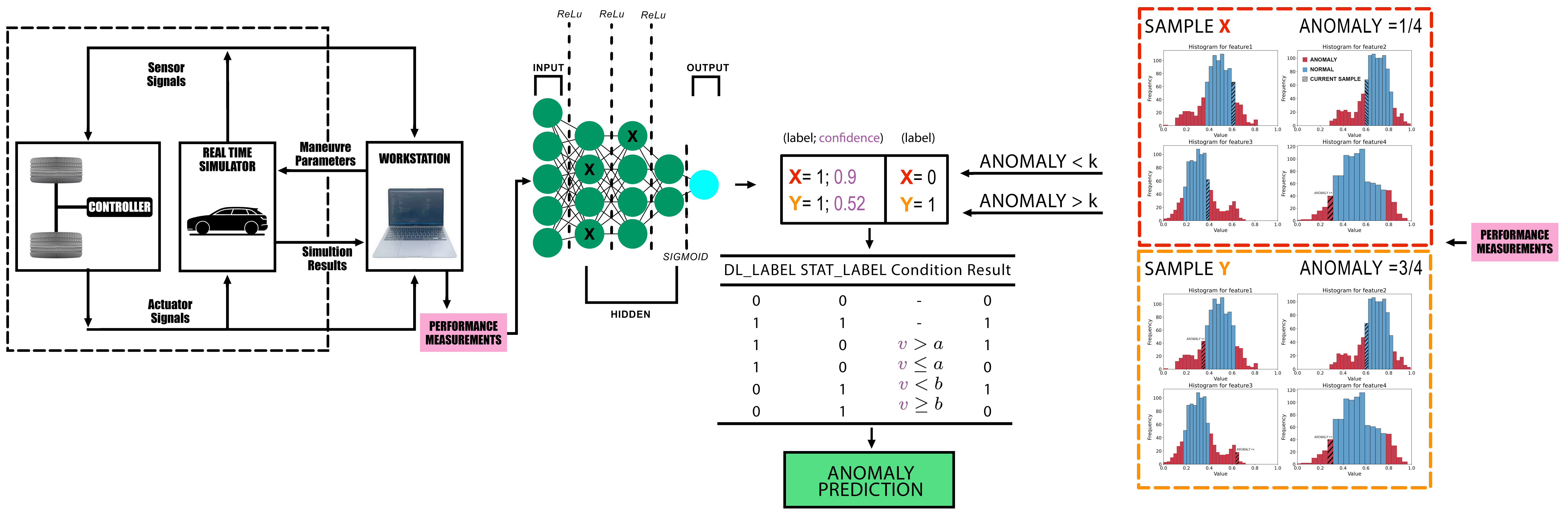}
    \caption{Architecture of the SAAD(Statistical Aggregated Anomaly Detection) framework, showcasing the integration of statistical anomaly detection with a Fully Connected Network (FCN) for enhanced anomaly detection in automotive systems. The architecture includes data preprocessing, artificial labeling, deep learning model training, and result aggregation to achieve improved accuracy and robustness. The effectiveness of SAAD is demonstrated with two samples, X and Y. The FCN initially predicts that both are anomalies, with sample X having high confidence and sample Y having low confidence. The statistical method, using \(k=2\), detects sample X as an anomaly and sample Y as not. Aggregation is applied to obtain the final answer, depending on the outputs of the models and the confidence of the deep learning one, combining the strengths of both methods.}
    \label{fig:main}
\end{figure*}

Anomaly detection is a critical task in various fields, particularly in the automotive industry, where it plays a significant role in ensuring the reliability and safety of systems. With the advent of advanced machine learning (ML) techniques, anomaly detection has seen substantial improvements in accuracy and efficiency. Machine learning models, especially deep learning architectures, have shown promising results in identifying abnormal patterns that could indicate potential faults or cyber-attacks in automotive systems \cite{jebur_review_2023, alsulami_symmetrical_2022}.

In the context of automotive systems, anomaly detection is crucial for maintaining the operational integrity of both traditional and autonomous vehicles. Autonomous vehicles, in particular, rely heavily on a multitude of sensors to navigate and operate safely. These sensors provide continuous streams of data that need to be monitored in real-time for any signs of anomalies that could compromise the vehicle's safety and functionality \cite{dixit2022anomaly, he2021machine}. Machine learning models have been employed to analyze these data streams, allowing for the early detection of issues and the implementation of corrective actions before any adverse events occur \cite{kim_comparative_2023}.

To validate and verify the performance of control systems under real-time conditions, Hardware-in-the-Loop (HIL) testing is an essential technique used in the industry. HIL testing involves integrating actual hardware components with simulated environments, providing a highly controlled and repeatable setting to assess the behavior of automotive systems. This method is particularly beneficial for anomaly detection, as it allows for the collection of high-fidelity data that can be used to train and test machine learning models under various scenarios \cite{matiecs2023detection}. Additionally, HIL testing helps mitigate the common issue of imbalanced data in anomaly detection datasets by enabling the generation of synthetic data to balance the dataset.

Evaluating anomaly detection algorithms often requires specific metrics tailored for time series data, as classical precision and recall metrics may not capture the nuanced performance characteristics required for detecting anomalies in continuous data streams \cite{huet_local_2022}. Advanced evaluation metrics consider the duration and timing of anomalies, providing a more robust assessment of an algorithm's performance in real-world conditions.

Recent studies have explored various machine learning approaches for anomaly detection in different domains. For instance, \cite{aslam_anomaly_2022} utilized explainable random forests to detect anomalies in oil wells, achieving high accuracy and demonstrating the effectiveness of explainable AI in critical applications. Another study \cite{kim_anomaly_2020} proposed a clustered deep one-class classification model for detecting anomalies in cyber-physical systems, showing significant performance improvements over traditional models. A survey on anomaly detection in IoT time-series data highlighted the unique challenges and proposed solutions for implementing these techniques in rapidly expanding fields like the Internet of Things \cite{cook_anomaly_2020}. Additionally, a comprehensive review on outlier detection in time series data provided a structured overview of unsupervised techniques and their applications \cite{blazquez-garcia_review_2021}.

The integration of ML in the industry has also yielded impressive results in structural health monitoring \cite{onchis_deep_2019, onchis_stable_2021}. Building on these advancements, recent studies have expanded the field towards hybrid-assisted models that leverage logical rules. These neuro-symbolic approaches combine deep learning with knowledge-based systems to enhance performance. Research has shown that these hybrid models, which incorporate domain-specific knowledge, significantly outperform standard deep learning models by providing more accurate and explainable results \cite{onchis_neuro-symbolic_2022, onchis_neuro-symbolic_2023}. Additionally, innovative methods such as GraphLIME have demonstrated significant improvements in model interpretability and performance in predictive tasks, further showcasing the potential of combining explainable AI techniques with deep learning \cite{costi_predictive_2024}. The interpretability aspect is highly valuable, as seen in \cite{hogea_fetril_2024}, where the insights were used to understand and solve the feature distribution issue that occurred with the generated pseudo-features. A comprehensive comparison in \cite{onchis_advantages_2022} highlights both the advantages of hybrid models and the challenges of creating and scaling them. These hybrid approaches, which synergize complementary components, form the cornerstone of the conceptual framework presented in this paper.

In this paper, we introduce Statistical Aggregated Anomaly Detection (SAAD), a novel anomaly detection method designed to effectively integrate statistical techniques for outlier detection with the performance of supervised machine learning models. Our approach specifically aims to enhance the accuracy and robustness of Fully Connected Networks (FCNs) in anomaly detection tasks. Utilizing real-world data collected from Hardware-in-the-Loop (HIL) environments, we demonstrate the effectiveness of SAAD in identifying anomalies with high precision and recall. Additionally, we illustrate how our statistical method can augment the performance of FCNs, rendering them more reliable for real-time applications in automotive systems

\section{Overview}
While our primary contribution is aggregation of the statistical method with a Fully Connected Network (FCN), the former is also a novel method. This method has been conceptualized by the first author and has been thoroughly adapted to an industrial setting with the help of the second author, proving to produce reliable results on its own. However, its full potential is realized when used in conjunction with a deep learning model. The deep learning model makes predictions and provides confidence scores. In cases where the confidence is low, the statistical method is applied to aggregate the predictions and produce a final, more accurate output.

The contributions of each author for the work presented in this paper are as follows:
\begin{itemize}
    \item \textbf{First author:} The first author conceptualized the statistical approach. He is also responsible for the creation of the artificial labeling, development of the aggregation function for labels, and conduction of the statistical method's experiments. 
    \item \textbf{Second author:} The second author played a crucial role in transitioning the theoretical method to an industrial application. Working with the baseline method provided by the third author. Introduced the idea of SAAD and helped showcase the results in this paper.
    \item \textbf{Third author:} Provided the thoroughly tested baseline model. The author also contributed by supplying and generating on demand the necessary data for extensive experiments. 
\end{itemize}

\section{Related Works}
The field of anomaly detection has garnered significant attention due to the increasing complexity and connectivity of modern systems. Several approaches have been proposed to address the challenges associated with identifying anomalies in various domains, with our particular interest in this paper being the combination of machine learning and statistical methods.

Intrusion Detection Systems (IDS) for networks, particularly the Controller Area Network (CAN) bus in automotive systems, have been extensively reviewed and developed. Lokman et al. \cite{lokman_intrusion_2019} provide a comprehensive overview of IDS implementation in the CAN bus system, discussing various detection approaches, deployment strategies, and technical challenges. They highlight the use of hybrid methods that combine frequency-based, machine learning-based, and statistical-based approaches for anomaly detection, emphasizing the importance of a holistic approach to securing networks.

Statistical learning methods have also been applied to anomaly detection in manufacturing environments, which share similarities with other systems in terms of operational variability and data characteristics. Pittino et al. \cite{pittino_automatic_2020} demonstrate the effectiveness of integrating statistical machine learning methods to control charts in detecting anomalies in production machines. Their hybrid approach, designed to function without continuous recalibration, achieves high recall and robustness, showcasing the potential for similar methods in various applications.

Deep learning techniques, particularly those involving Long Short-Term Memory (LSTM) networks, have shown promise in handling the complex and unpredictable nature of time-series data from multiple sensors. Malhotra et al. \cite{malhotra_lstm-based_2016} propose an LSTM-based Encoder-Decoder scheme for anomaly detection, which learns to reconstruct normal time-series behavior and uses reconstruction error to identify anomalies. This method, while primarily focused on machine learning, can benefit from statistical preprocessing to enhance its effectiveness across various types of time-series data, making it relevant for numerous applications.

The combination of statistical methods with machine learning has been explored in various fields, providing valuable insights for their application in anomaly detection. Holloway and Mengersen \cite{holloway_statistical_2018} review the use of statistical machine learning methods in remote sensing, highlighting their relevance for sustainable development goals. Similarly, Amin et al. \cite{amin_brain_2019} demonstrate the integration of statistical methods with machine learning for brain tumor detection, achieving high accuracy through a combination of denoising techniques and feature fusion.

ArunKumar et al. \cite{arunkumar_forecasting_2021} utilize Auto-Regressive Integrated Moving Average (ARIMA) and Seasonal Auto-Regressive Integrated Moving Average (SARIMA) models to forecast COVID-19 cases, underscoring the importance of combining statistical modeling with machine learning for accurate predictions. Banitalebi Dehkordi et al. \cite{banitalebi_dehkordi_ddos_2021} apply a hybrid approach to detect DDoS attacks in software-defined networks (SDN), integrating entropy-based methods with machine learning classifiers to enhance detection accuracy.

In the context of cryptocurrency price prediction, Khedr et al. \cite{khedr_cryptocurrency_2021} review traditional statistical and machine learning techniques, demonstrating the effectiveness of machine learning in handling the dynamic and non-seasonal nature of cryptocurrency data. Their work highlights the potential for similar hybrid approaches in anomaly detection.

A structured approach to anomaly detection for in-vehicle networks is presented by Muter et al. \cite{muter_structured_2010}. Their work focuses on a systematic methodology for identifying anomalies in vehicular communications, integrating statistical analysis with machine learning-based detection mechanisms to improve accuracy and robustness.

\textbf{Research Gap and Our Solution.}
Despite the significant advancements in anomaly detection, there remains a gap in effectively integrating statistical methods with machine learning models to achieve high accuracy and robustness, particularly in real-time applications. Many existing methods either rely heavily on machine learning techniques, which can be data-intensive and computationally expensive, or on statistical methods that may not fully capture the complexity of the data.

Our response to that is SAAD, which effectively demonstrates that standalone models, which perform similarly when used individually, benefit greatly from combining their results. By integrating statistical techniques with Fully Connected Networks (FCNs), SAAD enhances the detection of anomalies by leveraging the strengths of both approaches. Statistical preprocessing improves data quality, while dropout layers in FCNs prevent overfitting, leading to a more robust and accurate anomaly detection system. 

\section{Solution methodology}
In this section, we detail the comprehensive methodology employed in the development and validation of the SAAD framework. The methodology encompasses three main stages: (1) the artificial labeling procedure, (2) the development of the deep learning model, and (3) the aggregation of results from both methods to achieve enhanced anomaly detection performance. For a visual representation of the architecture of SAAD, we provide Figure \ref{fig:main}.

\subsection{Artificial labeling procedure} 
This stage primarily involves assigning artificial labels to dataset instances. This approach is rooted in the fact that most anomaly detection problems deal with unlabeled data. Typically, anomaly detection is an unsupervised learning task due to the difficulty in defining anomalies and the labor-intensive process of manually labeling data, which requires domain expert knowledge. Starting with the initial dataset, but excluding the target column, statistical methods are used to assign artificial or synthetic labels to the data. This method offers a different perspective compared to the supervised usage in deep learning, making their combination complementary. The term “artificial” indicates that the labels generated in this stage are not necessarily the ground-truth labels.

Anomalous measurements are identified for each feature of the dataset by assessing the occurrence probabilities of the values. The concept originates from the fact that anomalies are often not precisely defined; rather, they are described as instances that exhibit the common behavior \cite{big-survey}. Following this idea, in statistical terms, a value could be considered as being an anomaly if its occurrence probability is lower than a given threshold value. Since it is hard to work directly on high-dimensional data, each feature is examined separately. The occurrence probability of a given value is determined using the probability distribution function (PDF). Determining the exact probability density function (PDF) is often challenging, making it more practical to estimate the PDF using a histogram-based method. A histogram is created by splitting the initial collection of numerical values into groups called bins. Each bin has a defined lower and upper bound, which are used to determine where each value should be placed. The number and widths of the bins could be selected using different methods, but a proper width is determined using the Freedman–Diaconis rule \cite{Freedman1981}. The number of bins is inferred from the usage of bins with the same width.

\begin{figure}[!t]
\centering
\includegraphics[width=0.48\textwidth]{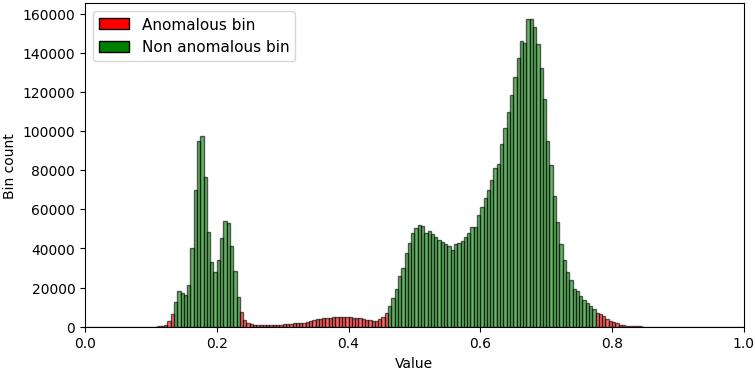}
\caption{Anomalous bins exemplification on histogram}
\label{anom-bins-img}
\end{figure}

The number of values from a given bin is called bin count. The relative frequency of a bin is calculated by dividing the bin count by the total count of all bins. Consider a bin being represented as \( b_i = \{ (a_{i-1}, a_i), \: c_i, \: m_i \}, \: i \geq 1 \): \( a_{i-1} \) and \( a_i \) are the bin's lower and upper bounds, \( c_i \) is the bin count and \( m_i \) is the relative frequency. Bins with high counts contain significantly more values, indicating that these values are more common within the initial collection. Consequently, the values in high-count bins have higher occurrence probabilities compared to those in low-count bins. A value that belongs to a bin with a low count could be considered as being an anomaly because it is not so common, i.e., it has a low number of occurrences. A bin \( b_i \) is called anomalous if its relative frequency value is lower than a given threshold value \( t \), i.e. \( m_i < t \). An exemplification of non-anomalous and anomalous bins on the histogram is presented in fig. \ref{anom-bins-img}. Further, a value \( x \in \mathbb{R} \) is considered as being an anomaly if it belongs to an anomalous bin.

The selection of anomalous bins is presented in Algorithm \ref{anomalous-bins-selection}. $B$ is a collection of collections: each collection $B_j$ contains the histogram bins for the values of feature $j$. The collection $A$ is used to retain the anomalous bins for each feature and is structured in the same way as $B$.
\begin{algorithm}[H]
\caption{Anomalous bins selection for all features}\label{anomalous-bins-selection}
\begin{algorithmic}
\REQUIRE dataset $D \subseteq  \mathbb{R}^{m \times n}$, $t \in [0, 1]$, bin width $k$ (optional, use 'auto' for Freedman–Diaconis rule)
\ENSURE $B = \{ B_0, B_1, ..., B_{n-1} \}$, $A = \{ A_0, A_1, ..., A_{n-1} \}$ 
\STATE

\STATE $A$, $B \gets [\textcolor{white}{x}], [\textcolor{white}{x}] $ \COMMENT {empty lists}
\STATE $j \gets 0$
\FOR[for each feature]{$j \gets 0$ to $n-1$} 
    \STATE $D^j \gets $ values of feature j from all instances
    \STATE $B_j \gets get\_bins(D^j,\, k)$
    \STATE add $B_j$ to $B$
    \STATE $i \gets 0$
    \STATE $A_j \gets [\textcolor{white}{x}] $ \COMMENT {empty list for anomalous bins}
    \FOR[for each bin]{$i \gets 0$ to $|B_j|-1$}
        \STATE $b_i \gets B_{j, i}$ \COMMENT {get current bin}
        \IF [compare relative freq. with threshold] {$m_i < t$}
            \STATE add $b_i$ to $A_j$
        \ENDIF
    \ENDFOR
    \STATE add $A_j$ to $A$
    \STATE
\ENDFOR
\STATE
\RETURN  $B$, $A$
\end{algorithmic}
\end{algorithm}

After the selection of anomalous bins for each feature, an instance-level anomaly is defined. An instance is considered an anomaly (label value 1) if it contains at least $k$ anomalous values (features). The procedure is presented in Algorithm \ref{cycle-artificial-labeling}.

\begin{algorithm}[h]
\caption{Artificial labeling procedure}\label{cycle-artificial-labeling}
\begin{algorithmic}
\REQUIRE instance $d \in \mathbb{R}^{1 \times n}$, $B = \{B_0, B_1, ..., B_{n-1}\}$, $A = \{A_0, A_1, ..., A_{n-1}\}$, $k < n$ 
\ENSURE 1 if $d$ is an (artificial) anomaly, 0 otherwise

\STATE $count \gets 0$

\FOR{$j \gets 0$ to $n-1$}
    \STATE $b \gets \texttt{bin\_of\_value}(d_j, B_j)$
    \IF {$b$ in $A_j$}
        \STATE $count \gets count + 1$
    \ENDIF
\ENDFOR

\IF { $count \geq k$}
    \RETURN {$1$}
\ENDIF
\RETURN {$0$}

\end{algorithmic}
\end{algorithm}

\subsection{Deep Learning Model}

The data preparation process is crucial for providing the model with clean and well-structured input data. Initially, data is loaded from CSV files located in a specified directory, generating file paths and reading the CSV files. During loading, the data undergoes cleaning to handle missing or undefined values, ensuring the dataset is free from any irregularities that could affect model training. Relevant features, or signals, are then selected from the dataset based on the specified configuration, focusing on the necessary columns for model training. Categorical features have been encoded and subsequently, the data is labeled based on the presence of anomalies, with a binary label (0 or 1) assigned to each data point to indicate whether it is an anomaly.

After labeling, the data is scaled using a standard scaler to normalize the feature values. This normalization step helps in speeding up the convergence during training by ensuring that all features contribute equally to the learning process.

The dense network architecture comprises several fully connected (dense) layers with dropout layers between them to prevent overfitting. The input layer uses the selected relevant features and feeds them to the hidden layers, which consist of multiple dense layers with ReLU activation functions and L2 regularization to mitigate overfitting. The output layer is a single neuron with a sigmoid activation function to produce a binary output indicating the presence or absence of an anomaly.

The model is compiled using the Adam optimizer, a binary cross entropy loss function, and various evaluation metrics including binary accuracy, precision, recall, and custom F-beta scores.

During inference, the model processes new input data in the same manner, outputting a prediction along with a confidence score. The prediction is binary, indicating whether the input data point is an anomaly, while the confidence score reflects the model's certainty about this prediction. The confidence score is particularly important in applying SAAD, as the aggregation of results from the deep learning model and the statistical model is based on the confidence from the former.

\subsection{Aggregated solution}
The results provided by the artificial labeling procedure and the deep learning model are aggregated to create a new set of labels. These final results are compared with the ground-truth labels to compute the performance metrics for the proposed approach. For an instance $d \in D$, consider \texttt{STAT\_LABEL} as the label (prediction result) provided by the artificial labeling procedure and \texttt{DL\_LABEL} as the label provided by the deep learning model for the evaluation of $d$. For an instance $d$ evaluated by the deep learning model, the outcome value \texttt{DL\_LABEL} (either 1 or 0) depends on the value $v$, computed at the end of the evaluation using the sigmoid function. The value $v$ is further used by the deep learning model to decide the outcome for the evaluated instance $d$: if, $v > 0.5$ then, the model classifies $d$ as an anomaly (label value 1), otherwise as a non-anomalous instance (label value 0). The aggregation function decides the final label value using the values of \texttt{STAT\_LABEL}, \texttt{DL\_LABEL}, and $v$. The decision cases are described below:

\begin{table}[H]
\centering
\begin{tabular}{ccccl}
\toprule
\textbf{\texttt{STAT\_LABEL}} & \textbf{\texttt{DL\_LABEL}} & \textbf{Condition} & \textbf{Result} \\ 
\midrule
0 & 0 & - & 0 \\
1 & 1 & - & 1 \\
0 & 1 & $v > a$ & 1 \\
0 & 1 & $v \leq a$ & 0 \\
1 & 0 & $v < b$ & 1 \\
1 & 0 & $v \geq b$ & 0 \\
\bottomrule
\end{tabular}
\vspace{0.1cm}
\caption{Decision rules for label aggregation}
\label{tab:aggregation-rules}
\end{table}

\begin{table*}[!ht]
\centering
\caption{Accuracy Scores Obtained on Aggregated Results (in \%)}
\label{acc-score-final}
\begin{tabular}{@{}ccccccccccccccccccc@{}}
\toprule
\textbf{$b$ \textbackslash $a$} & \textbf{0.51} & \textbf{0.52} & \textbf{0.53} & \textbf{0.54} & \textbf{0.55} & \textbf{0.60} & \textbf{0.65} & \textbf{0.70} & \textbf{0.75} & \textbf{0.80} & \textbf{0.85} & \textbf{0.90} & \textbf{0.95} & \textbf{0.96} & \textbf{0.97} & \textbf{0.98} & \textbf{0.99} & \textbf{1} \\ \midrule
\textbf{0.00} & 54.0 & 54.3 & 54.6 & 54.8 & 54.9 & 56.2 & 59.1 & 59.9 & 61.1 & 62.2 & 63.4 & 65.4 & 68.3 & 73.0 & 77.1 & 77.1 & 77.1 & 77.1 \\
\textbf{0.01} & 61.8 & 62.0 & 62.3 & 62.5 & 62.7 & 63.9 & 66.9 & 67.6 & 68.8 & 69.9 & 71.2 & 73.2 & 76.0 & 80.7 & 84.8 & 84.8 & 84.8 & 84.8 \\
\textbf{0.02} & 62.1 & 62.4 & 62.7 & 62.9 & 63.1 & 64.3 & 67.2 & 68.0 & 69.2 & 70.3 & 71.5 & 73.5 & 76.4 & 81.1 & 85.2 & 85.2 & 85.2 & 85.2 \\
\textbf{0.03} & 62.4 & 62.7 & 62.9 & 63.1 & 63.3 & 64.6 & 67.5 & 68.3 & 69.4 & 70.6 & 71.8 & 73.8 & 76.6 & 81.3 & 85.4 & 85.4 & 85.4 & 85.4 \\
\textbf{0.04} & 62.6 & 62.8 & 63.1 & 63.3 & 63.5 & 64.7 & 67.6 & 68.4 & 69.6 & 70.7 & 71.9 & 74.0 & 76.8 & 81.5 & 85.6 & 85.6 & 85.6 & 85.6 \\
\textbf{0.05} & 62.7 & 63.0 & 63.2 & 63.4 & 63.6 & 64.8 & 67.7 & 68.5 & 69.7 & 70.8 & 72.1 & 74.1 & 76.9 & 81.6 & 85.7 & 85.7 & 85.7 & 85.7 \\ 
\textbf{0.10} & 63.8 & 64.0 & 64.3 & 64.5 & 64.7 & 65.9 & 68.8 & 69.6 & 70.7 & 71.1 & 72.4 & 74.4 & 77.2 & 81.9 & 86.0 & 86.0 & 86.0 & 86.0 \\ 
\textbf{0.15} & 64.4 & 64.7 & 64.9 & 65.1 & 65.3 & 66.5 & 69.4 & 70.2 & 71.3 & 71.4 & 72.6 & 74.6 & 77.4 & 82.1 & 86.2 & 86.2 & 86.2 & 86.2 \\ 
\textbf{0.20} & 64.8 & 65.0 & 65.3 & 65.5 & 65.7 & 66.9 & 69.8 & 70.6 & 71.7 & 71.6 & 72.8 & 74.8 & 77.6 & 82.3 & 86.4 & 86.4 & 86.4 & 86.4 \\ 
\textbf{0.25} & 65.1 & 65.4 & 65.6 & 65.8 & 66.0 & 67.2 & 70.1 & 70.9 & 72.0 & 71.7 & 72.9 & 75.0 & 77.8 & 82.5 & 86.6 & 86.6 & 86.6 & 86.6 \\ 
\textbf{0.30} & 65.4 & 65.7 & 65.9 & 66.1 & 66.3 & 67.5 & 70.4 & 71.2 & 72.3 & 71.8 & 73.1 & 75.1 & 77.9 & 82.6 & 86.7 & 86.7 & 86.7 & 86.7 \\ 
\textbf{0.35} & 65.8 & 66.0 & 66.3 & 66.5 & 66.7 & 67.9 & 70.8 & 71.6 & 72.7 & 72.0 & 73.3 & 75.3 & 78.2 & 82.9 & 86.9 & 86.9 & 86.9 & 86.9 \\
\textbf{0.40} & 66.2 & 66.4 & 66.7 & 66.9 & 67.1 & 68.3 & 71.2 & 72.0 & 73.1 & 72.1 & 73.3 & 75.3 & 78.2 & 82.9 & 86.9 & 86.9 & 86.9 & 86.9 \\ 
\textbf{0.45} & 66.6 & 66.8 & 67.1 & 67.3 & 67.5 & 68.7 & 71.6 & 72.4 & 73.5 & 72.2 & 73.4 & 75.4 & 78.3 & 83.0 & 87.1 & 87.1 & 87.1 & 87.1 \\ 
\textbf{0.46} & 66.9 & 67.1 & 67.4 & 67.6 & 67.8 & 69.0 & 71.9 & 72.7 & 73.8 & 72.6 & 73.8 & 75.8 & 78.7 & 83.4 & 87.5 & 87.5 & 87.5 & 87.5 \\
\textbf{0.47} & 67.1 & 67.4 & 67.6 & 67.8 & 68.0 & 69.2 & 72.1 & 72.9 & 74.0 & 72.8 & 74.0 & 76.0 & 78.9 & 83.6 & 87.7 & 87.7 & 87.7 & 87.7 \\
\textbf{0.48} & 67.3 & 67.6 & 67.8 & 68.0 & 68.2 & 69.4 & 72.3 & 73.1 & 74.2 & 73.0 & 74.2 & 76.2 & 79.1 & 83.8 & 87.9 & 87.9 & 87.9 & 87.9 \\
\textbf{0.49} & 67.6 & 67.8 & 68.1 & 68.3 & 68.5 & 69.7 & 72.6 & 73.4 & 74.5 & 73.2 & 74.4 & 76.4 & 79.3 & 84.0 & 88.1 & 88.1 & 88.1 & 88.1 \\
\textbf{0.50} & 67.8 & 68.0 & 68.3 & 68.5 & 68.7 & 69.9 & 72.8 & 73.6 & 74.7 & 73.4 & 74.6 & 76.6 & 79.5 & 84.2 & 88.3 & 88.3 & 88.3 & 88.3 \\
\bottomrule
\end{tabular}
\end{table*}

\begin{table*}[!ht]
\centering
\caption{F1 Scores Obtained on Aggregated Results}
\label{f1-score-final}
\begin{tabular}{@{}ccccccccccccccccccc@{}}
\toprule
\textbf{$b$ \textbackslash $a$} & \textbf{0.51} & \textbf{0.52} & \textbf{0.53} & \textbf{0.54} & \textbf{0.55} & \textbf{0.60} & \textbf{0.65} & \textbf{0.70} & \textbf{0.75} & \textbf{0.80} & \textbf{0.85} & \textbf{0.90} & \textbf{0.95} & \textbf{0.96} & \textbf{0.97} & \textbf{0.98} & \textbf{0.99} & \textbf{1} \\ \midrule
\textbf{0.00} & 0.52 & 0.53 & 0.53 & 0.54 & 0.54 & 0.56 & 0.60 & 0.61 & 0.63 & 0.65 & 0.66 & 0.69 & 0.72 & 0.78 & 0.82 & 0.82 & 0.82 & 0.82 \\
\textbf{0.01} & 0.63 & 0.64 & 0.64 & 0.64 & 0.65 & 0.66 & 0.70 & 0.71 & 0.72 & 0.74 & 0.75 & 0.78 & 0.81 & 0.85 & 0.89 & 0.89 & 0.89 & 0.89 \\
\textbf{0.02} & 0.64 & 0.64 & 0.65 & 0.65 & 0.65 & 0.67 & 0.70 & 0.71 & 0.73 & 0.74 & 0.76 & 0.78 & 0.81 & 0.85 & 0.89 & 0.89 & 0.89 & 0.89 \\
\textbf{0.03} & 0.64 & 0.65 & 0.65 & 0.65 & 0.66 & 0.67 & 0.71 & 0.72 & 0.73 & 0.75 & 0.76 & 0.78 & 0.81 & 0.86 & 0.89 & 0.89 & 0.89 & 0.89 \\
\textbf{0.04} & 0.65 & 0.65 & 0.65 & 0.66 & 0.66 & 0.67 & 0.71 & 0.72 & 0.73 & 0.75 & 0.76 & 0.78 & 0.81 & 0.86 & 0.89 & 0.89 & 0.89 & 0.89 \\
\textbf{0.05} & 0.65 & 0.65 & 0.65 & 0.66 & 0.66 & 0.68 & 0.71 & 0.72 & 0.74 & 0.75 & 0.76 & 0.79 & 0.81 & 0.86 & 0.89 & 0.89 & 0.89 & 0.89 \\ 
\textbf{0.10} & 0.66 & 0.66 & 0.67 & 0.67 & 0.67 & 0.69 & 0.72 & 0.73 & 0.75 & 0.75 & 0.77 & 0.79 & 0.82 & 0.86 & 0.90 & 0.90 & 0.90 & 0.90 \\ 
\textbf{0.15} & 0.67 & 0.67 & 0.67 & 0.68 & 0.68 & 0.69 & 0.73 & 0.74 & 0.75 & 0.75 & 0.77 & 0.79 & 0.82 & 0.86 & 0.90 & 0.90 & 0.90 & 0.90 \\ 
\textbf{0.20} & 0.67 & 0.67 & 0.68 & 0.68 & 0.68 & 0.70 & 0.73 & 0.74 & 0.76 & 0.76 & 0.77 & 0.79 & 0.82 & 0.87 & 0.90 & 0.90 & 0.90 & 0.90 \\ 
\textbf{0.25} & 0.68 & 0.68 & 0.68 & 0.68 & 0.69 & 0.70 & 0.74 & 0.75 & 0.76 & 0.76 & 0.77 & 0.79 & 0.82 & 0.87 & 0.90 & 0.90 & 0.90 & 0.90 \\ 
\textbf{0.30} & 0.68 & 0.68 & 0.68 & 0.69 & 0.69 & 0.71 & 0.74 & 0.75 & 0.77 & 0.76 & 0.77 & 0.80 & 0.82 & 0.87 & 0.90 & 0.90 & 0.90 & 0.90 \\ 
\textbf{0.35} & 0.68 & 0.68 & 0.69 & 0.69 & 0.69 & 0.71 & 0.74 & 0.76 & 0.77 & 0.76 & 0.78 & 0.80 & 0.83 & 0.87 & 0.91 & 0.91 & 0.91 & 0.91 \\
\textbf{0.40} & 0.68 & 0.69 & 0.69 & 0.69 & 0.70 & 0.71 & 0.75 & 0.76 & 0.77 & 0.76 & 0.78 & 0.80 & 0.83 & 0.87 & 0.91 & 0.91 & 0.91 & 0.91 \\ 
\textbf{0.45} & 0.69 & 0.69 & 0.69 & 0.70 & 0.70 & 0.72 & 0.75 & 0.76 & 0.78 & 0.76 & 0.78 & 0.80 & 0.83 & 0.87 & 0.91 & 0.91 & 0.91 & 0.91 \\ 
\textbf{0.46} & 0.69 & 0.69 & 0.70 & 0.70 & 0.70 & 0.72 & 0.75 & 0.76 & 0.78 & 0.77 & 0.78 & 0.80 & 0.83 & 0.87 & 0.91 & 0.91 & 0.91 & 0.91 \\
\textbf{0.47} & 0.69 & 0.70 & 0.70 & 0.70 & 0.70 & 0.72 & 0.76 & 0.77 & 0.78 & 0.77 & 0.78 & 0.80 & 0.83 & 0.88 & 0.91 & 0.91 & 0.91 & 0.91 \\
\textbf{0.48} & 0.70 & 0.70 & 0.70 & 0.70 & 0.71 & 0.72 & 0.76 & 0.77 & 0.78 & 0.77 & 0.79 & 0.81 & 0.84 & 0.88 & 0.92 & 0.92 & 0.92 & 0.92 \\
\textbf{0.49} & 0.70 & 0.70 & 0.71 & 0.71 & 0.71 & 0.73 & 0.76 & 0.77 & 0.79 & 0.77 & 0.79 & 0.81 & 0.84 & 0.88 & 0.92 & 0.92 & 0.92 & 0.92 \\
\textbf{0.50} & 0.70 & 0.71 & 0.71 & 0.71 & 0.71 & 0.73 & 0.77 & 0.78 & 0.79 & 0.78 & 0.79 & 0.81 & 0.84 & 0.89 & 0.92 & 0.92 & 0.92 & 0.92 \\
\bottomrule
\end{tabular}
\end{table*}

\section{Experiment}

\begin{figure*}[!ht]
    \centering
    \includegraphics[width = 0.98\textwidth]{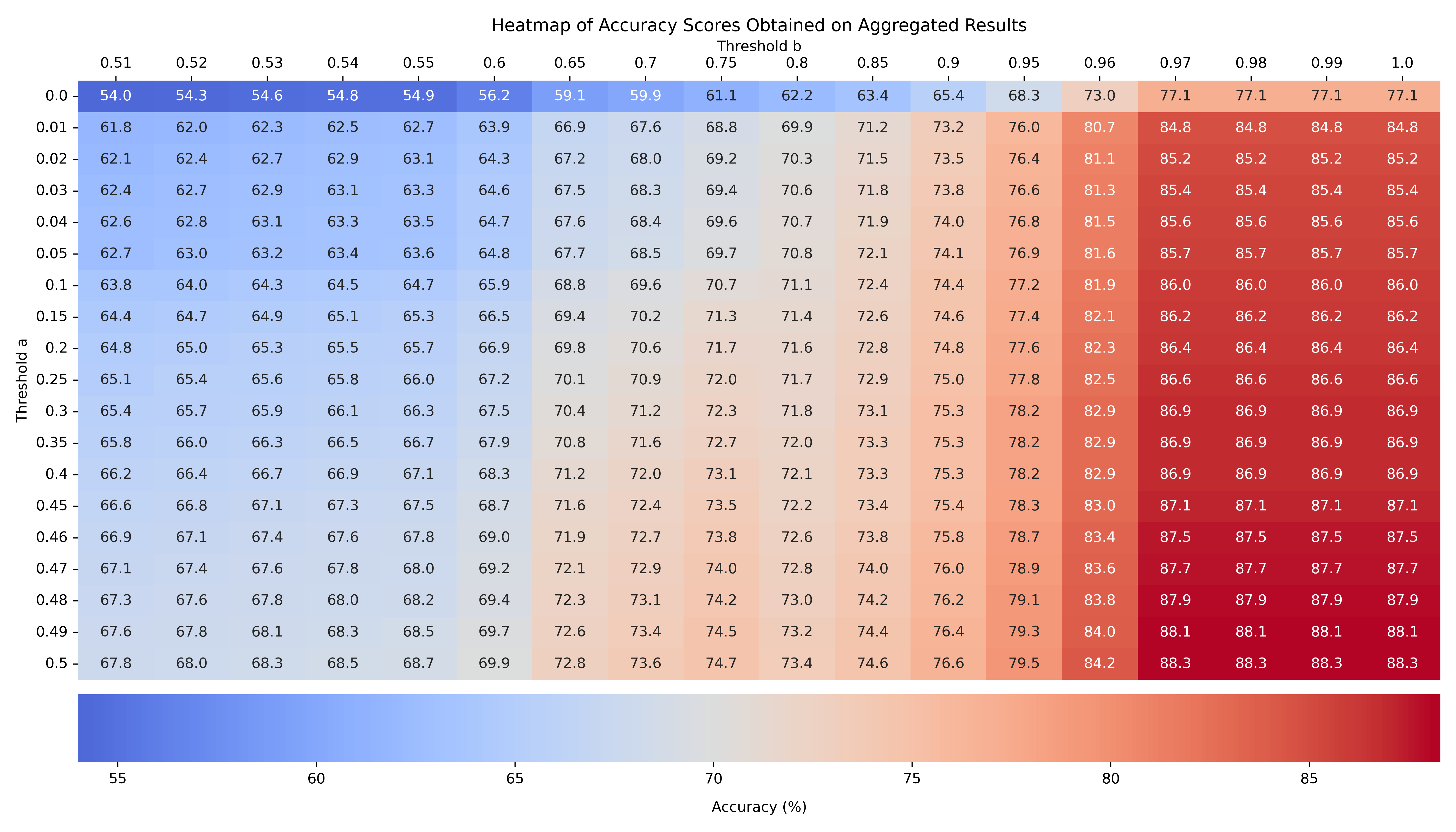}
    \caption{Heatmap of Accuracy Scores Obtained on Aggregated Results. The heatmap visualizes the accuracy percentages across different thresholds \(a\) and \(b\). The color intensity represents the accuracy, with warmer colors indicating higher accuracy scores. The pattern shows that accuracy generally increases as both thresholds \(a\) and \(b\) increase, with the highest accuracy observed in the lower right corner of the heatmap, where both thresholds are at their maximum values. This indicates a positive correlation between higher threshold values and improved accuracy.}
    \label{fig:enter-label}
\end{figure*}

\subsection{Data Acquisition}
The data acquisition process for this study focuses on measuring and collecting data from the Hardware-in-the-Loop (HIL) environment, specifically tailored for underbrake scenarios. This is also presented in the Figure \ref{fig:main}. The workbench setup includes several components to simulate and capture the necessary data:

\textbf{Simulation Workbench.}
The simulation workbench connects a real-time simulation environment to the physical Electronic Control Unit (ECU) of a vehicle. The simulation environment encompasses detailed models of various vehicle systems such as the engine, transmission, suspension, and steering. These models interact with sensor readings, control signals, and actuator commands to mimic real-world driving conditions.

The HIL setup provides inputs to the ECU, which then processes these inputs using its control algorithms. The vehicle's response, including feedback from sensors such as wheel speed, brake pressure, and brake pad temperature is sent back to the simulation environment. A benefit of this feedback loop is that it allows engineers to test and validate the performance of the vehicle's braking system and identify any unwanted behavior in the ECU's algorithms without the need for physical testing.

\textbf{Data Acquisition System (DAQ).}
The DAQ system is a critical component of the workbench, designed to measure and collect high-fidelity data from the HIL environment. The DAQ system comprises the following elements:
\begin{itemize}
    \item \textbf{Sensors}: Devices that measure various parameters during the simulation, including temperature, position, voltage, and other relevant metrics.
    \item \textbf{Data Acquisition Hardware}: This hardware interfaces the sensors with the computer and typically includes an analog-to-digital converter to transform the analog signals from the sensors into digital signals that can be processed by the computer.
    \item \textbf{Software}: The DAQ software is used to configure the hardware, process the collected data, and display the results. It includes drivers necessary for data acquisition and tools for data analysis.
    \item \textbf{Computer System}: A robust computer system is utilized to run the DAQ software and store the collected data for subsequent analysis.
\end{itemize}

\textbf{Data Collection Process.}
The raw data is represented by measurements obtained from the ECU within the HIL environment. Each measurement consists of thousands of signals sampled at 10ms intervals, with each signal potentially representing a unique feature.

To gather these measurements, various vehicle maneuvers were simulated to reproduce underbrake scenarios. Key variables for these maneuvers included vehicle speed (ranging from 0 to 150 km/h) and brake pedal pressure (ranging from 5\% to 100\%). These maneuvers ensured a diverse and representative dataset, which is essential for training robust machine learning models.


\textbf{Experimental Setup.}
For this study, a total of 150 vehicle maneuvers were simulated for the training set to capture a wide range of underbrake scenarios. Additionally,  distinct maneuvers were created for the testing set, featuring multiple underbrake activations and varying speeds. This approach helps to evaluate the performance of the anomaly detection algorithms and prevents overfitting.

By carefully designing the data acquisition and preprocessing steps, we ensure that the machine learning models trained on this data can make accurate predictions and decisions, ultimately improving the reliability of automotive systems. 

\subsection{Experiment methodology}
As proof of concept, the proposed method was used to detect anomalies on the data collected from a HIL environment: as mentioned in previous section, the used data represent measurements taken from different usages of braking function of vehicles on different situations; the used datasets contain 18 features and the target class.




\textbf{Models parameters configuration.} In the case of artificial labeling procedure, the used parameters are: number of minimum anomalous values per instance $k = 3$ and anomalous bin relative frequency threshold $t = 0.05$. The artificial labeling method works on an unsupervised way: it assigns labels just by usages of statistical methods, without using ground truth labels. The used parameters values were chosen such that, the percentage of artificial anomalous labels to be closer to the percentage of ground-truth anomalous labels. 

For the deep learning model, to further enhance the model's performance, hyperparameter optimization is applied using Bayesian optimization with Keras Tuner. The optimizer defines a search space for several key hyperparameters, including the number of units in each dense layer, learning rate, dropout rates, and regularization parameters. 

The optimization process iteratively evaluates different combinations of these hyperparameters by training the model on the training data and validating it on a separate validation set. Early stopping is employed to prevent overfitting, halting training when the model's performance on the validation set ceases to improve. The best-performing model configuration is then selected based on the evaluation metrics.

\subsection{Experiments results}
The artificial labeling method alone provided an accuracy score of 72.1\% on the testing data. The deep learning model alone provided on accuracy score of 71.5\% on the testing data. The labels provided by each method were aggregated according to the presented strategy. The new labels were compared with ground-truth labels and accuracy and F1 score metrics were used to measure the performances: the results are presented in the table \ref{acc-score-final} and \ref{f1-score-final}.

\begin{figure}[!htbp]
    \centering
    \includegraphics[width = 0.49\textwidth]{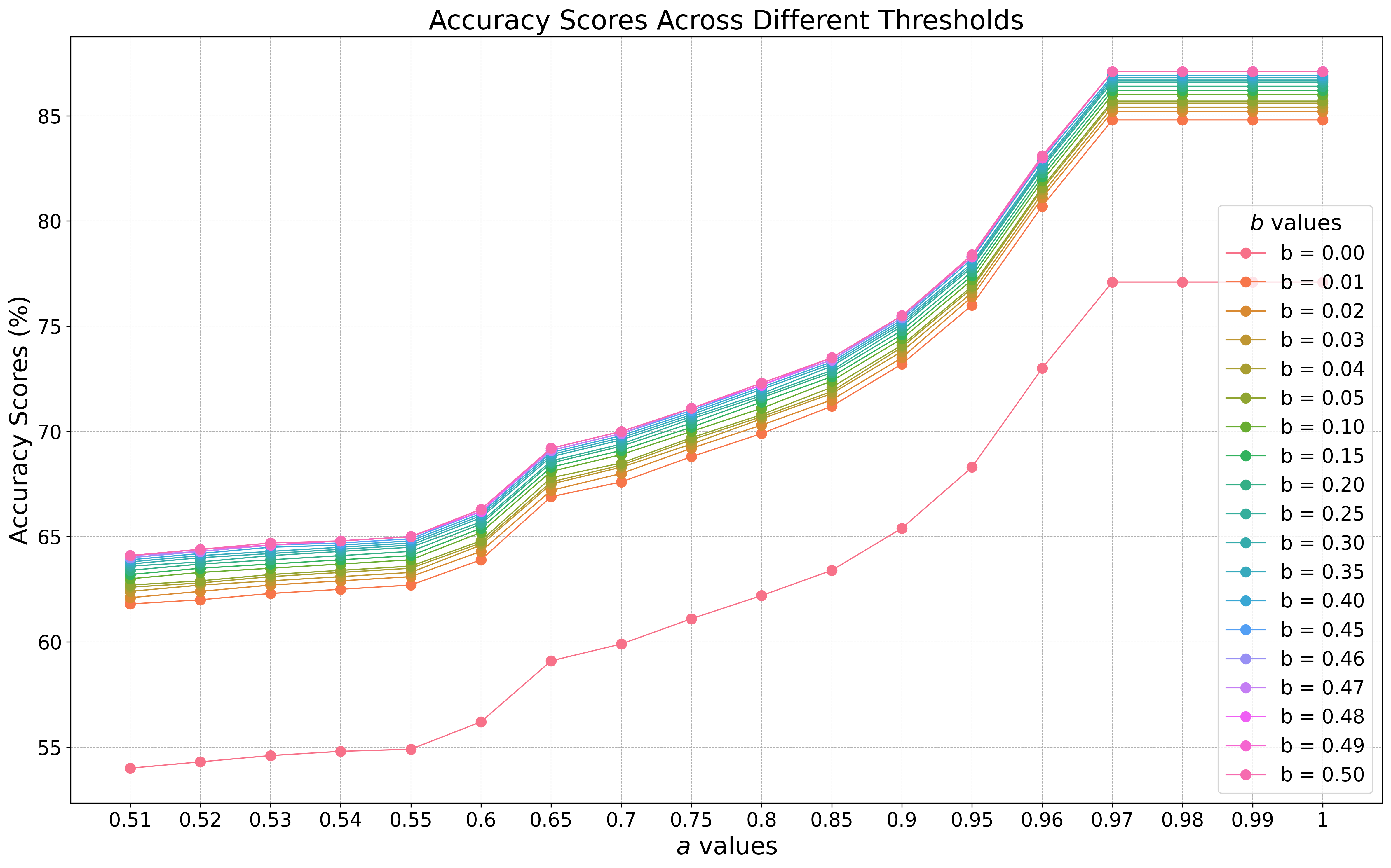}
    \caption{Evolution of accuracy at different threshold values.}
    \label{fig:annex-evolution}
\end{figure}

As can be observed from results from the tables \ref{acc-score-final} and \ref{f1-score-final}, the proposed approach that combines artificial labeling method with deep learning model provide higher scores than each method alone. Both parameters $a$ and $b$ from aggregation function have an important role on the quality of the results. Higher $a$ and $b$ allow the used approach to provide better results: labels obtained with artificial labeling method together with higher $a$ and $b$ values enforce the aggregation function to select high confidence results provided by deep learning model. The experimental results for different pairs of $a$, $b$ values can also be seen in Figure \ref{fig:annex-evolution}.

\subsection{Results interpretation}
As presented in the table \ref{tab:aggregation-rules}, the aggregation approach provided higher performances, in comparison with both models alone. As an initial explanation for the linear / uniform increase in the accuracy and F1 score values, this trend is likely related to the use of small steps in the values used in the tuning for aggregation parameters $a$ and $b$. As explanations related to the higher performances provided by the aggregation strategy, firstly for the case when both models provide the same label, the aggregated label is well understood. 

When the deep learning model classifies an instance as an anomaly (label value 1) and the statistical approach classifies it as a normal instance (label value 0), we rely on the deep learning model's result only if it provides a high enough confidence for the given prediction. For example, if $v = 0.55$ this value is close to 0.5, which is the upper bound for what is considered normal. Thus, we use $v>a$ to ensure that the prediction quality is sufficient to consider the evaluated instance as an anomaly. By increasing the value of $a$, we request a higher confidence $v$ (expect progressively better result), effectively raising the threshold for accepting its predictions. This higher expectation leads to a decrease in the number of instances labeled as anomalies (value 1) provided by the deep learning model. If $a = 1$, the condition  $v > 1$ holds nowhere, meaning all instances will be labeled as normal (value 0), relying solely on the statistical approach's labels.

When the deep learning model classifies an instance as non-anomalous (label value 0) and the statistical approach classifies it as an anomaly (label value 1), we rely on the deep learning model if its confidence value $v$ is low enough. A $v$ value lower than 0.5 but still close to it indicates a low confidence level for the evaluated instance to be non-anomalous, so we enforce prediction quality using a constraint rule based on parameter $b$: $v < b$. In the case of the deep learning model, an instance $d$ is considered as being normal if $v \in [0, 0.5]$; thus, if $b = 0.5$, the condition $v<0.5$ always holds, thus all such cases receive the label 0 (not anomaly), according to the value provided by the deep learning model.

To reiterate, for $a=1$ and $b=0.5$, when the two models provide distinct labels for an evaluated instance, related to the outcomes provided by the aggregation method, all anomaly labels (value 1) are coming from the statistical approach, while all non-anomaly labels (value 0) come from the deep learning model. The best performance is provided by the aggregation function using the parameters $a=1$ and $b=0.5$ indicating that the statistical approach performs better at predicting anomalies, whereas the deep learning model excels at predicting non-anomalous instances. An explanation for previous observation relates to the nature of the used models. The deep learning model, due to the unbalanced nature of the dataset (with many more normal instances compared to anomalies), may have a bias favoring the prediction of normal instances over anomalies, leading to better performance in predicting normal instances. Similarly, the statistical approach's superiority in predicting anomalies is likely due to its mode of operation: it determines the anomalous or non-anomalous state of an evaluated instance by examining feature values and checking for low-frequency values. This approach uses several raw thresholds to decide what is an anomaly and what is normal. The statistical approach's outcome can heavily rely on just a few feature values of the evaluated instance. For example, with $k=4$, means that if the evaluated instance contains at least 4 anomalous values, the instance is classified as an anomaly, even if the rest of the attributes have common values (with respect to the PDF). Therefore, the statistical approach is stricter regarding the values of an instance, making it easier to declare an instance as anomalous: a sufficient number of anomalous values can override the common (non-anomalous) values in the rest of the attributes.

\section{Conclusion}
In this paper, we introduced SAAD, a novel statistical anomaly detection method with promising results for automotive systems, validated through extensive testing using real-world data collected from a Hardware-in-the-Loop (HIL) environment. Our main contribution is merging a novel statistical method with deep learning, taking the best of both approaches and achieving higher accuracy with the joint operation than with standalone models.

The results of our experiments demonstrate the efficacy of our proposed method. The standalone statistical method achieved an accuracy of 72.1\% on the testing data, while the deep learning model alone provided an accuracy of 71.5\%. When combined, the aggregated results yielded an accuracy of up to 88.3\% and an F1 score of 0.921.

The integration of statistical techniques with machine learning models highlights the potential for enhancing existing anomaly detection frameworks, making them more reliable and effective in real-world applications.

Future work will focus on further refining the statistical method, exploring adaptive threshold techniques for multivariate time series data, and expanding the application scope to other critical automotive systems. This will help improve the robustness and scalability of our approach, ensuring its effectiveness across diverse and complex scenarios, while also extending its applicability beyond automotive systems.

\section*{Acknowledgment}
We extend our gratitude to Prof. Darian Onchis for his invaluable guidance and support during the planning stages of this work, and for shaping the direction of our research. His initial works on hybrid models were crucial in recognizing their potential and applicability in the industry.

\end{document}